\documentclass[journal]{IEEEtran}
   \usepackage{graphicx}
\usepackage{url}

\begin{document}
%
\title{Interaction Histories and Short Term Memory: Enactive Development of Turn-taking Behaviors in a Childlike Humanoid Robot}
%
%
%

\author{\IEEEauthorblockN{Frank Broz, Chrystopher L. Nehaniv, Hatice Kose-Bagci, and Kerstin Dautenhahn} \\
\IEEEauthorblockA{Adaptive Systems Research Group\\
School of Computer Science\\
University of Hertfordshire, UK\\
Email: f.broz, c.l.nehaniv, h.kose-bagci, k.dautenhahn @herts.ac.uk}}

\maketitle

\begin{abstract}
In this article, an enactive architecture is described that allows a humanoid robot to learn to compose simple actions into turn-taking behaviors while playing interaction games with a human partner. The robot's action choices are reinforced by social feedback from the human in the form of visual attention and measures of behavioral synchronization.  We demonstrate that the system can acquire and switch between behaviors learned through interaction based on social feedback from the human partner. The role of reinforcement based on a short term memory of the interaction is experimentally investigated. Results indicate that feedback based only on the immediate state is insufficient to learn certain turn-taking behaviors. Therefore some history of the interaction must be considered in the acquisition of turn-taking, which can be efficiently handled through the use of short term memory.

\end{abstract}


%
\IEEEpeerreviewmaketitle

\section{Introduction}
%
%
%
%
\IEEEPARstart{S}{ocial}  interaction plays a vital role in a child's development. At an early age, 
children acquire 
basic social skills, such as the ability to read
social gaze and  how to play simple games that involve turn-taking, that serve as a scaffold for more sophisticated forms of socially-mediated learning. Children learn these skills through
interaction with their caretakers, motivated by intrinsic social drives that cause them
to seek out social engagement as a fundamentally rewarding experience. Research on the innately social nature of child development has highlighted the primary role intersubjective experience of an infant with a carer (e.g.~Trevarthen \cite{Trevarthen79}, Trevarthen~\cite{trevarthen01intrinsic}) and of contingency and communicative imitation (e.g.~Nadel et al.~\cite{nadel99expectancies}, Kaye~\cite{Kaye82}). 

It is desirable to have robots learn to interact in a similar manner, both to gain 
insight into how social skills may develop and to achieve the goal of more natural
human-robot interaction. This paper describes a system for the learning of behavior
sequences based on rewards arising from social cues, allowing a childlike humanoid robot to engage a human interaction partner 
in a social interaction game. This system builds on earlier behavior learning work using interaction histories in our lab by Mirza and colleagues~(\cite{mirza07grounded,mirza08developing}) as part of the  RobotCub  project \cite{RobotCubWebsite} on the iCub humanoid developed by our consortium. The goals of this project include providing an open-platform for cognitive systems research and our work focuses on the ontogeny of behaviour in such systems.\footnote{The software developed and used in this work is available open-source at \url{http://sourceforge.net/projects/robotcub/}.} 
The work follows an enactive paradigm (e.g.\ \cite{varela91,dautenhahn96remembering,mirza07grounded,vernon10,hutto12}) that has been identified as a promising framework for scaling behaviour-based approaches to fully learning and developing cognitive robotic systems (these concepts are expanded on in Section~\ref{sec:enactive}). 
Autonomy, embodiment, emergence, and experience play the key roles via situated interaction (including social interaction) in  the course of the  ontogeny of cognition
\cite{varela91,dautenhahn96remembering,vernon10}.   Temporal grounding via the remembering and reapplication of experience, as well as the social dimension of interaction, are a central themes addressed in the interaction history architecture (IHA), which has been further extended here.  

  In this article, the results of a human-robot interaction experiment are presented that demonstrates how this enactive learning architecture supports the acquisition of behaviors, how learned behaviors can be switched between based on social cues from the human partner, and how the inclusion of a short term memory in the architecture supports the acquisition of more complex turn-taking behaviors.


The desire for social contact is a basic drive that motivates behavior in humans from a very young age. Studies by Johnson and Morton indicate that infants are able to recognize and are attracted to human faces~\cite{johnson91biology}.
Juscyck has demonstrated that they are also especially attracted to the sound of the human voice as compared to other noises~\cite{juscyck92developing}. Work by Farroni and colleagues shows that they are capable of recognizing when they are the object of another's gaze, and look longer at the faces of those looking at them~\cite{farroni02eye}. Nadel's studies demonstrate that they are also able to recognize contingent behavior and prefer it to noncontingent behavior~\cite{nadel99expectancies}. And from the very earliest stages of life, many developmental psychologists argue that they begin to engage in simple interactions with aspects of turn-taking
(e.g.~imitating the facial gestures of others as studied by Meltzoff and Moore~\cite{meltzoff89imitation}, in intersubjective development (Trevarthen and Aitken~\cite{trevarthen01infant}),  and contingency in interaction and the development of communicative imitation~(Kaye~\cite{Kaye82}, Nadel et al.~\cite{Nadel99,nadel99expectancies})).

Infants thus seem to find many aspects of social interaction innately rewarding. This suggests the possibility of using these features to provide reward to a developmental learning system.
Such a system would enable robot behavior to be shaped through interaction in a way that is inspired by the infant-caretaker relationship and engage humans in a manner that could assist in scaffolding the development of behavior. Some possible sources of reward for such a system (based on the examples given above) could be:

\begin{itemize}
\item Presence of a person
\item Vocalization/prosody
\item Visual attention
\item Contingency of another's actions
\item Synchronization of both partners' actions
\end{itemize}

This list is organized roughly in order from more immediate perceptual feedback to qualities of an interaction that appear to require some history of the relationship between of one's own and another's behavior over time. In this study, feedback based on immediate perception and on a history of interaction is used to reinforce behaviors in the learning architecture. A simple computational model of short term memory is introduced in order to transiently capture some of this recent history and facilitate the learning of social interaction behaviors.

\section{Background}

 Research in both infant development and in robotics informs our approach to the developmental learning architecture and learning scenario presented in this paper. The behaviors that are investigated in the context  of this work on social development are gaze and turn-taking. The importance of short-term memory in social interaction is explored, and relevant research on the development of short-term memory in infants is presented. Finally, the use of child-like robots in the study of human-robot interaction and developmental learning is discussed.

\subsection{Social Behaviors}
\label{sec_socialdev}
\subsubsection{Gaze}

Gaze is an especially powerful social cue. It is also one that becomes significant at an 
early developmental stage; a study by Hains and Muir indicates that even young infants are responsive to others' gaze direction 
\cite{hains96adult}.  Corkum and Moore have demonstrated that infants are able to follow eye movements from around 18 months of age \cite{corkum95development}. 
The simplest response to a gaze cue is the recognition of having another's visual attention. In a review of the literature on observer response to gaze, Frischen, Bayliss and Tipper suggest that this recognition  is the basis of and developmental precursor to more complex gaze behaviors such as joint attention~\cite{frischen07gaze}.  This ability is crucial in a social context, as it provides valuable feedback about whether a nearby person is looking at (and therefore ready to interact with) someone or whether they are attending to something else. There is also some evidence that social gaze is inherently rewarding to people. A recent fMRI study by Schilbach and colleagues of a joint attention task found activation of the ventral striatum, an area of the brain involved in processing reward \cite{schilbach10eyes}.

\subsubsection{Turn-taking and Games}

Turn-taking plays a fundamental role in regulating human-human social
interaction and communication from an early age. Trevarthen describes how turn-taking ``proto-conversations" between infants and caretakers set the stage for more complex social interaction and later language learning~\cite{trevarthen01infant}. Elias, Hayes and Broerse studied how mothers control and shape early vocal play with infants into turn-taking interactions~\cite{elias86maternal}. Rutter's early studies on turn-taking conclude that carers appear to train infants how to engage in conversational turn-taking through interaction~\cite{rutter87turn-taking}.  Ross and Lollis suggest that the ability to engage in turn-taking games is a skill that children begin to develop early in life~\cite{ross87communication}.
Ratner and Bruner claim that the regularity due to restriction on possible actions and repetition of simple social turn-taking games allow infants to learn to predict the order of events and increase their agency in the interaction by reversing roles during turn-taking~\cite{ratner78games}.  Turn-taking dynamics do not depend on the behavior of a single individual but emerge from the interaction between partners. In human-human interaction and communication, the role switch between ``leader" and ``follower" is not determined by external sources. Humans manage when to start and stop their turns in social interactions based on various factors including the context and purpose of the interaction, feedback from social interaction partners, and emotional and motivational factors.

Peek-a-boo is a simple social game played with small children where an adult blocks the child's view to the adult's face with their hands and then lowers them to reveal their face while saying ``peek-a-boo". This game has often been a scenario used for the study of social learning in developmental psychology. Bruner and Sherwood studied mothers playing peek-a-boo with their children in order to investigate how parents teach the rules and structure of interaction behavior to infants \cite{bruner76peek}. Peek-a-boo has also been used by Rochat et al. in order to study infants' ability to recognize regular or irregular patterns of action in familiar interactions~\cite{rochat99emerging}.
Gustafson, Green, and West observed changes in peek-a-boo play as infants' mode of play transitioned from passive to active during development \cite{gustafson79infant}. As children come to understand the nature of the game, it can take on a turn-taking form, with the adult and child alternating who is the one hiding their face. 

Drumming has also been used to study social interaction in children. Kirschner and Tomasello demonstrated that children participating in social drumming tasks (where they drummed with a present human partner rather than a disembodied beat) were more likely to adapt their drumming to that of their interaction partner~\cite{kirschner09joint}. A recent study by Kleinspehn-Ammerlahn and colleagues showed that the ability to synchronize with a partner while drumming was age-dependent, improving as children approach adulthood~\cite{kleinspehn11dyadic}. A review of the literature by Accordino, Comer and Heller shows that drums and other percussion instruments are commonly used in music therapy for autistic children that seeks to improve childrens' social skills~\cite{accordino07searching}. A study by Kim, Wigram and Gold on music therapy that included drum play found improvements in joint attention and turn-taking behaviors in children with autism~\cite{kim08effects}.

\subsection{Short Term Memory}

There is increasing experimental evidence that people rely on short term memory in order to evaluate others' social behavior. A study by Phillips, Tunstall and Shelley demonstrates that  decoding certain kinds of nonverbal social cues requires the use of short term memory \cite{phillips07exploring}. As found by Fuster and Alexander, the prefrontal cortex is widely believed to be involved in storing and processing short term memory \cite{fuster71neuron}.
Recent work by Chan and Downing provides evidence that the prefrontal cortex's response to human faces (as opposed to that of other parts of the brain) is focussed on the eyes and may be involved in gaze processing \cite{chan11faces}. FMRI studies by Kuzmanovic and colleagues suggest that the prefrontal cortex is involved in making judgements about social gaze related to gaze duration~\cite{kuzmanovic09duration}.  The prefrontal cortex has also been observed to be involved in the processing of temporal information in short term memory tasks in primates, such as the recognition of the frequency pattern of a simulus in work by Romo, Brody and Lemus \cite{romo99neuronal} or remembering the temporal ordering of events for an action in work by Ninokura, Mushiake, and Tanji \cite{ninokura03representation}.

In child development, researchers are in agreement that infants are born with limited working memory capacity that significantly improves by the latter half of their first year. Most experimental research on short term memory in infants focuses on the identification of objects, which makes it somewhat difficult to find direct implications for open-ended, movement-based play interactions (this can be seen in the focus of the research covered in the recent book on infant memory by Oakes and Bauer~\cite{oakes07short}). But some researchers, such as Chen and Cowan, approach the measurement of short-term memory in terms of time rather than symbolic chunks \cite{chen05chunk, cowan05working}. Studies of working memory in infants performed by Diamond and Doar and also by Schwartz and Reznick indicate that they can perform short-term memory based tasks requiring a memory of up to 10 seconds by around 9 to 12 months of age~\cite{diamond89performance,schwartz99measuring}.

Baddeley and Hitch originally proposed a model of working memory that included separate information stores for short-term storage of auditory and visio-spatial information, with a central executive to process this information~\cite{baddeley74working}. 
An episodic buffer was later added by Baddeley to this model 
in his theory of working memory~\cite{baddeley00episodic} 
to reflect the fact that working memory appears to form associations across different sensory perceptions and make use of this associated information 
.  Baddeley's episodic buffer operates on multimodal input and preserves temporal information, allowing the analysis of the sequential history of recent experience. The other two submodules in his proposed model of working memory deal with the specialized processing auditory and visual and spatial information. The central executive manages switching between tasks and selective attention, as well as coordinating the transfer of information between the other parts of the system.
The form that our short term memory module takes (described in Section~\ref{sec:STM}), operating over a temporal sequence of past sensorimotor data, is related to Baddeley's model.

\subsection{Social and Developmental Robotics} 

The idea of using a childlike appearance to encourage people to engage in social interaction with a robot in a way that mimics the scaffolding behaviors that support human learning is an approach that has been adopted by a number of researchers in developmental robotics.
The robot Affetto was designed to be infant-like in order to study the role of the attachment relationship that forms between infants and their caretakers in child development~\cite{ishihara11realistic}. The CB2 robot was designed to learn social skills related to joint attention, communication, and empathy from humans~\cite{asada09cognitive}. The iCub robot, the robot used in this experiment, is another childlike robot designed as a platform for research in cogntive systems and developmental robotics, including the acquisition of social competencies learned from humans~\cite{tsakarakis07icub}.
Other social robots designed for human-robot interaction, such as 
Simon~\cite{chao11simon}, 
   Nexi~\cite{nexi},
  Robovie~\cite{imai02robovie},
 and the Nao~\cite{nao}
are designed to be childlike in either size, facial appearance, or both. However, for these robots, the design decision was made primarily to make the robot appealing to interact with rather than specifically to support developmental approaches to learning through interaction. 
Prior to availability of the iCub, the forerunner of the present architecture \cite{mirza08developing} was implemented on KASPAR, a minimally expressive, childlike humanoid developed in our lab~\cite{Kaspar}.

The iCub is a 53 degree-of-freedom (dof), toddler-sized humanoid robot. It has a color camera embedded in the center of each eye. Its facial expression can be changed by turning on and off arrays of LEDs underneath the translucent cover of its face to create different shapes for the mouth and eyebrows. The iCub's control software, including the architectures described in this article, is open source and the source code is available for download as part of the Robotcub project software repository.
The modules that make up its control system communicate using the YARP middleware for interprocess communication~\cite{metta06yarp}.

Roboticists such as Breazeal and Scassellati and child development researchers such as Meltzoff have proposed using robots to study the development of social cognition ~\cite{scassellati02theory, breazeal05learning, meltzoff07like}. However, these proposed approaches have focussed on the development of theory of mind and its role in imitation learning. In our work, we are exploring 
other mechanisms that support social engagement. 
An infant does not need to understand the motivations or beliefs of another person to derive enjoyment from their presence and contingent interaction.
Rather than focussing on intentionality, this system focuses on the coordination of behavior and social feedback as mechanisms for scaffolding turn-taking social interactions.

\section{Related Work}

The study of the emergence of turn-taking has vital implications in many areas like robot-assisted therapy, especially in relation to children with autism, where turn-taking games have been
used to engage the children in social interaction~\cite{robins04effects}. In many turn-taking studies, the focus has been on evaluating turn-taking quality rather than learning how to take turns from the human. Koizumi et al.\ found that delays and gaps in a robot's time of response led to negative evaluations by human participants~\cite{koizumi06preliminary}. Ito and Tani studied joint attention and turn-taking in an imitation game played with the humanoid robot QRIO~\cite{ito04joint}. In this game the human participants tried to find the action patterns, which were learned by QRIO previously, by moving synchronously with the robot. The robot KISMET used social cues for regulating turn-taking in non-verbal interactions with people~\cite{breazeal03towards}. Turn-taking between KISMET and humans emerged from the robot's internal drives and its perceptions of cues from its interaction partner, but the robot did not learn. In a recent study by Chao and colleagues, humans played a turn-taking imitation game with a humanoid robot \cite{chao11simon}. The purpose of this study was to collect data in order to investigate the qualities of successful turn-taking in multimodal interaction.  Kose-Bagci and colleagues studied emergent role-switching during turn-taking using a drumming interaction with a childlike humanoid robot that compared various rules for initiating and ending turns~\cite{kose-bagci10effects}.  The aim of this research was not to produce psychologically plausible models of human turn-taking behavior but to employ simple, minimal generative mechanisms to create different robotic turn-taking strategies based on social cues.  The robot dynamically adjusted its turn length based on the human's last turn but did not learn to sequence the drumming actions themselves.

Peek-a-boo has been used previously as the interaction scenario to demonstrate developmentally motivated learning algorithms for robots. 
Work by Mirza et al.\ demonstrated a robot learning to play peek-a-boo through interaction with a human using an earlier version of the interaction history architecture (to be described in the next section)~\cite{mirza08developing}. 
Ogino et al.\ proposed and implemented a developmental learning system that used reward to select when to transfer information from an agent's short term memory to long term memory. The test scenario used to demonstrate this learning system was a task of recognizing correct or incorrect performance of peek-a-boo based on an experiment with human infants~\cite{ogino07acquiring}. 
In both of these studies, peek-a-boo was only behavior acquired so the systems did not demonstrate the learning of more than one interaction task within a single scenario.

Kuriyama and colleagues have recently conducted research on robot learning of social games through interaction~\cite{kuriyama10learning}. They focus on different aspects of interaction learning than the work described in this article, using causality analysis to learn interaction rules by discovering correlations in the low-level sensorimotor data stream. Our work differs in that it makes use of social cues (such as human gaze directed at the robot) that are fundamental to human interaction and representative of the perceptual capabilities of a human infant (though they may appear more high-level from a data-processing standpoint). Also, the social games we seek to learn in this study are specified to the human interaction partner, who attempts to teach the robot a set of specific skills.\footnote{As our work is here is a cognitive systems development project, and not a human-robot interaction study, using an skilled experimenter in the role of the human interaction partner is appropriate to assess and demonstrate the capabilities of the system. In particular, its exhibiting particular behaviours with this interaction partner shows that it is possible for the system to achieve them in principle. And, its failure to exhibit certain behaviours under particular settings even with this trained partner suggests that is unlikely to achieve them with more na\"{i}ve interaction partners. This allows us to argue that certain components of the architecture are either sufficient or necessary to achieve certain behaviours using this enactive cognitive architecture. This, in turn, lays the groundwork, for further studies that could be carried out with large groups of human participants, as would be appropriate in future HRI studies.    }  The robot has only very minimal social and interactive motivations for what constitutes success in its behavior in the form of rewards for social engagment. The focus of the learning is on composing actions to produce behavior and coordinating the robot's behavior with the human's
in the context of interaction and social cues from the human, and could be used for acquiring various behaviors, including in principle ones not shown in our experiments.

\subsection{The Interaction History Architecture}
\label{sec:IHA}

This research extends past work on the iCub using the Interaction History Architecture 
(IHA).  To avoid confusion, the original  system will be referred to as IHA and the modified system that is the focus of this article will be referred to as the Extended Interaction History Architecture (EIHA). EIHA differs from the previous version in that the types of sensor input available to the robot has been expanded and a short term memory module has been added to the architecture.

IHA is a system for learning behavior sequences for interaction based
on grounded sensorimotor histories. While the robot acts, it builds up a memory of past
``experiences'' (distributions of sensors, encoders, and internal variables based on a 
short-term temporal window). Each experience is associated with the action the 
robot was executing when it occurred, as well as a reward value based on properties
of the experience. These experiences are organized for the purpose of recall using Crutchfield's information metric as a distance measure~\cite{crutchfield90information}. The information metric (which the authors will also refer to as the information distance) is the sum of the conditional entropies of two random variables each conditioned on the other. Unlike other popular ways of comparing distributions (such as mutual information or Kullback-Leibler divergence
) it meets the conditions necessary to define a metric space. Given two random variables, the information distance between them is defined as the sum of the conditional entropy of each variable conditioned on the other.

\begin{equation}
d(X,Y) = H(X|Y) + H(Y|X)
\end{equation}

In IHA, the random variables are sensor inputs or actuator readings whose values are measured over a fixed window of time that is defined as a parameter of the system. These values define a distribution for each sensor or actuator that is assumed to be stationary over the duration of time defined by the window. This vector of variables provides an operationalization of temporally extended experience for the robot. As the robot acts in the world, the informational profile (probability distribution) of this sensorimotor experience (including internal variables) as well as reward is collected as an `experience' in a growing body of such experiences that is structured as a dynamically changing metric space comprised of such experiences. The pairwise sum of the information distances between the variables making up the current experience vector and a past experience vector are calculated and used as a distance metric. As the robot acts, the most similar past sensorimotor experience  to its current state is found, and new actions are probabilistically selected based on their reward value, which is dynamically changing based on the interaction.  For a full description of the architecture, see prior publications by Mirza et al.~\cite{mirza07grounded, mirza08developing}. 

This architecture is inspired by earlier work by Dautenhahn and Christaller proposing cognitive architectures for embodied social agents in which the remembering and behavior are tightly interwined via internal structural changes and agent-environment interaction dynamics~\cite{dautenhahn96remembering}, and is informed by the enactive paradigm for cognitive architectures (cf.\ \cite{varela91,vernon10}). We view both the embodiment of the system and its embedding in a social context to be critical aspects of the system's cognition.  Our focus on learning through social interactions 
rather than an agent learning exclusively by exploring the world on its own is based on the hypothesis that social interaction was the catalyst to the development of human intelligence and therefore is a promising key domain for the development of robot intelligence~\cite{dautenhahn95getting}.

While the original version of IHA implements a model of remembering in which sensorimotor experiential data  may be recalled in a way that allows experiences to be compared to one another, it lacks certain characteristics that are useful, possibly even necessary for learning about social interaction. IHA makes no use of how recently experiences have occurred. Also, while experiences are themselves temporally extended, it is only over a short duration temporal horizon, and only single experiences rather than sequences of experiences are compared. 

Additionally, in IHA all rewards are calculated based solely on the current sensor data. For many tasks, especially interaction tasks, this is not sufficient to determine whether the phenomena that should be reinforced is occurring unless the temporal horizon of experiences has been chosen to match the task \cite{MirzaPhD}. In many cases, it is necessary to examine a sequential history of the sensor data for relationships between an agent's actions and those of the agent with which it interacts. We hypothesize that fluid turn-taking requires attention to the recent history of both one's own and the other's actions in order to anticipate and prepare for the shift in roles. In light of this, EIHA incorporates a short term memory over the recent history of sensor data relevant to the regulation of turn-taking (to be described in Section~\ref{sec:STM}).

\section{Architecture Design and Implementation}

EIHA is intended to support the robot developing different socially communicative, 
scaffolded behaviors in the course of temporally extended social interactions with 
humans by making use of social drives and its own firsthand experience of 
sensorimotor flow during social interaction dynamics. The robot comes to associate sequences of simple actions, such as waving an arm or hitting the drum, executed under certain conditions with successful interaction based on its past experience. 

In order to demonstrate these concepts, rewards based on social drives are designed
to influence the development of behavior in  an 
open-ended face-to-face interaction game between the iCub and a human. The human interaction partner interacts with the robot and may provide it with 
positive social feedback by directing their face and gaze toward the robot, as well as by engaging in drum play. The rewards reflecting these social drives for human attention
and synchronized turn-taking may be based on the current state of the robot's sensorimotor
experience or on the recent history of experience transiently kept in its short term memory.
The robot uses this feedback to acquire behavior that leads to sustained 
interaction with the human.

The system described in this article focuses on the use of visual attention as a form of social feedback, and on the learning of turn-taking behavior to explore issues of contingency and synchronization in interaction. Two forms of non-verbal turn-taking are specified as behaviors to be learned by the iCub through interaction with a human, peek-a-boo and drumming. 

\begin{figure*}[!t]
\centering
\includegraphics[width=.8\textwidth]{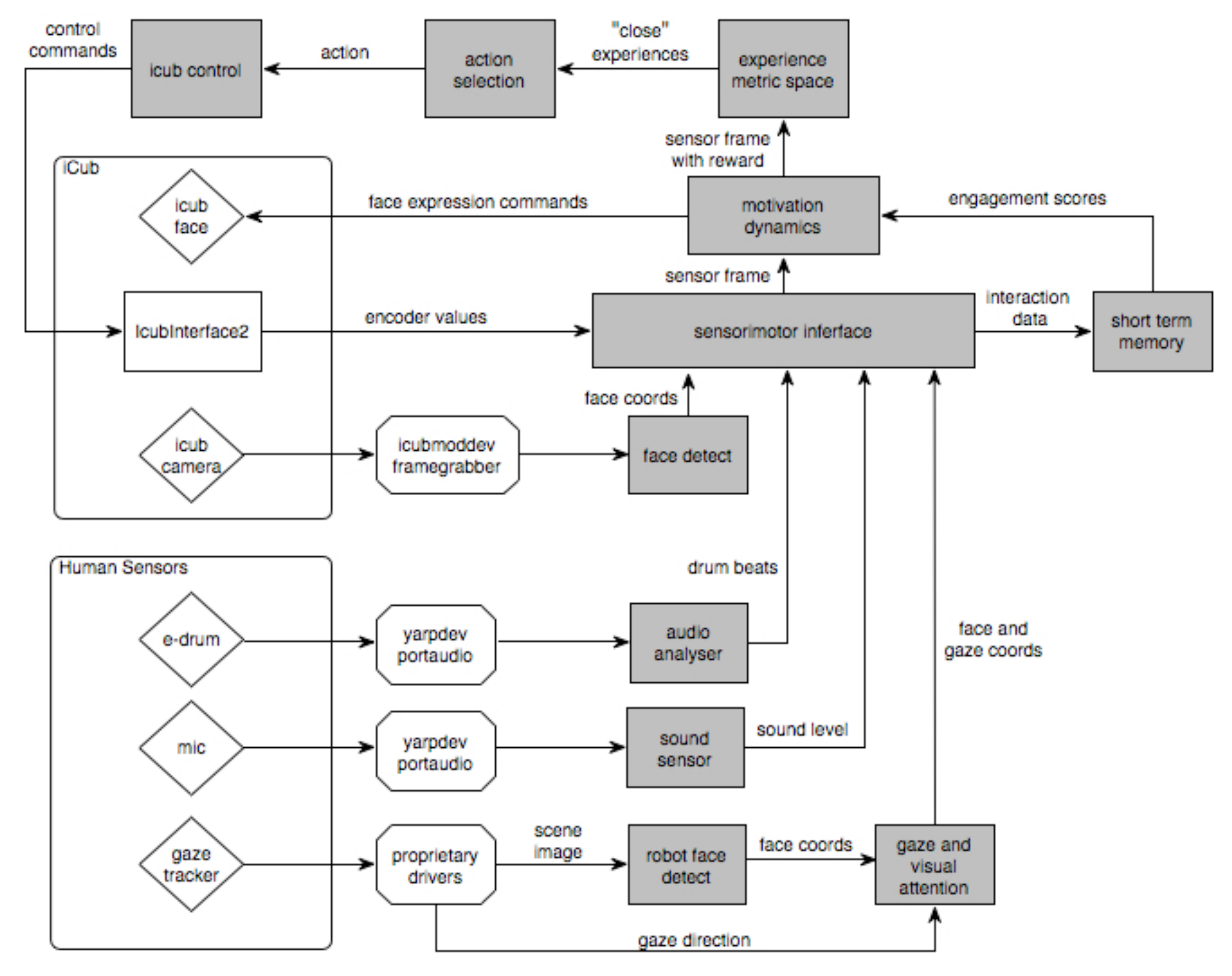}
\caption{Diagram of the Extended Interaction History Architecture (EIHA), sensor input, and robot controller.}
\label{fig_iha}
\end{figure*}

The implementation of EIHA used for learning the social interaction game is shown in Figure~\ref{fig_iha}.  Details of the sensorimotor data used as input to the learning system and the actions available to the robot for this scenario are described. The role of the short term memory module in assigning reward and the design of the rewards based on immediate sensor input and short term memory are explained. Finally, a description of how this learning architecture supports switching between learned behaviors during interaction is given.

\subsection{Sensorimotor data}
 
 The sensorimotor data  whose joint distribution over temporally extended time intervals
 make up an experience are a collection of data variables describing the robot's own joint positions and both raw and processed sensor input. Continuous-valued data were discretized using 8 equally-sized bins over the range of each variable. The raw camera image is captured as a very low resolution (64 pixel) intensity image. High level information from processed sensor data is included as binary variables reporting the presence of a face, the detection of a drum beat by the human drummer, and whether the human is currently looking at the robot's face. A full list of data variables is given in Table~\ref{tab_sensorimotor}.
 
 The system takes sensor input from the robot's eye camera, an external mic (this input source was not used for the experiment described in this article), an electronic drum, and an ASL MobileEye gaze-tracking system~\cite{mobileeye}. The wearable gaze tracking system included a head-mounted camera that records the scene in front of the person. The system outputs the person's gaze target within this scene image in pixel coordinates. 
 
Certain sensor input was processed to provide high-level information about the person's current activities. Face detection was used to locate the human interaction partner's face in the camera image from the robot's eye, and a binary variable reported whether or not a person was currently visible to the robot. Face detection was also applied to the gaze tracker's scene camera image to locate the robot's face, compare the location to the person's gaze direction, and report whether the person was looking at the robot as a binary variable (face detected in Table~\ref{tab_sensorimotor}). The Haar-wavelet-based face detection algorithm implementation in OpenCV was used. The audio stream from the electronic drum was filtered to extract drum-beats using the method implemented for the drum-mate system described in the appendix of Kose-Bagci et al.~\cite{kose-bagci10effects}.  Note also that variables referring to the ``engagement score" for the drumming and hide-and-seek tasks are also a part of the robot's experience. Their inclusion links the immediate experience of the robot to its recent history of observed behavior as interpreted by the short term memory module. This relationship will be further explained in Section~\ref{sec:STM}.

 \begin{table}
\caption{Data variables that make up an experience in the experience metric space.}
\label{tab_sensorimotor}
\centering
\begin{tabular}{|l|r|}
\hline
Data & \# of variables \\ \hline
 left eye intensity image & 64  \\
 head position & 3  \\
 eyes position & 3  \\
 left arm position & 7 \\
 left hand position& 9  \\
 right arm position & 7  \\
 right hand position& 9  \\
 face detected & 1  \\
 beat detected & 1\\
 visual attention score & 1  \\
 drum engagement score &  1\\
 hide engagement score & 1\\ \hline
Total & 107\\ \hline
\end{tabular}
\end{table}

\subsection{Actions and Behaviors}

 \begin{table*}[t]
\caption{Low level actions that the robot selects among during execution (ending pose shown).}
\label{table_actions}
\centering
\begin{tabular}{cccccc}
 \includegraphics[width=.14\textwidth]{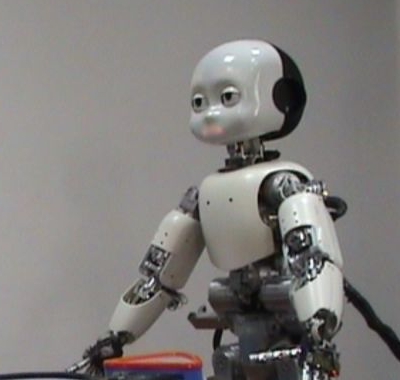}&
   \includegraphics[width=.14\textwidth]{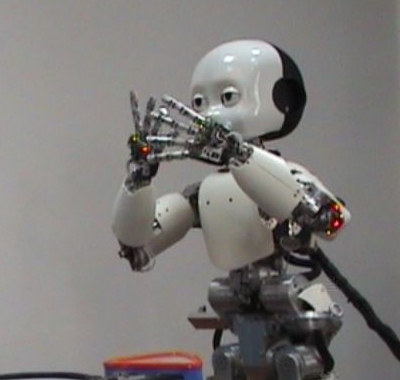}&
     \includegraphics[width=.14\textwidth]{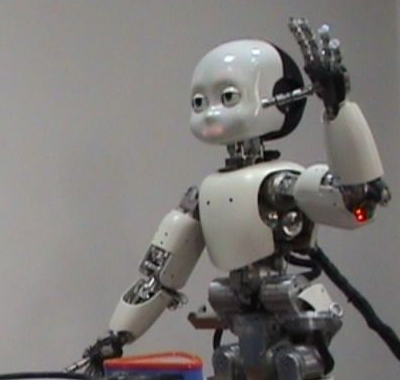}&
       \includegraphics[width=.14\textwidth]{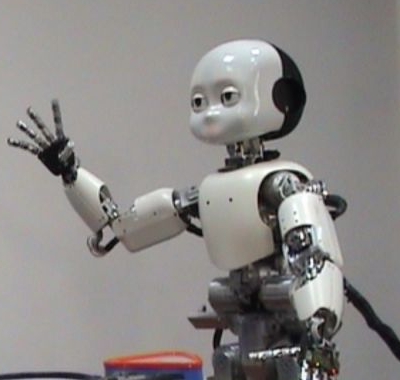}&
         \includegraphics[width=.14\textwidth]{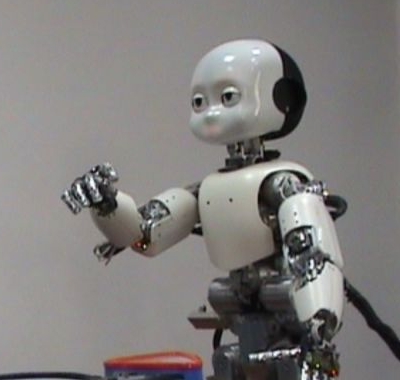}&
           \includegraphics[width=.14\textwidth]{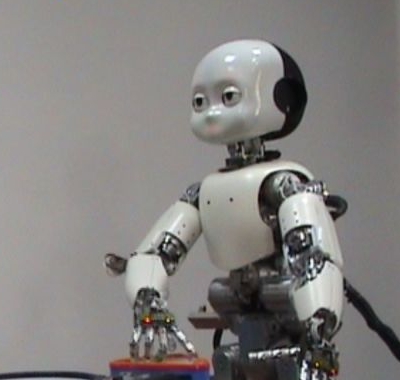}\\
  right-arm-down,&
   hide-face&
   right-arm-up, left-arm-up&
  right-arm-wave,& 
   start-drum& 
   drum-hit\\
      left-arm-down, home&&&left-arm-wave&&\\
\end{tabular}
\end{table*}


The robot's actions are preprogrammed sequences of joint positions and velocities that make up a recognizable unit of motion. The full list of 10 actions plus a short-duration no-op action available to the robot is given in Table~\ref{table_actions} (note that actions which have the same end point or are reflections of one another are represented by one image). The robot discovers the reward values of actions for a particular experience during execution. EIHA probabilistically chooses actions associated with high reward. Over time, the robot learns behavior sequences by experiencing sequences of high reward actions and associating them with a sequence of experiences, and selecting those actions again when those remembered experiences are similar to its current experience. It is important to note that while the low-level actions may have some recognizable ``purpose" to the human interactor, the behaviors that are to be learned are made up of sequences of these actions. For example, the peek-a-boo behavior is made up of ``hide-face", ``home-position", and a single turn of drumming is made up of ``start-drum", an arbitrary number of ``drum-hit" actions, and then ``right-arm-down" or ``home-position". These sequences may also have an arbitrary number of ``no-op" actions interspersed throughout them, affecting their timing.

\subsection{Short Term Memory Module}
\label{sec:STM}

In addition to a dynamic memory of sensorimotor experience and associated rewards, 
it is also useful to have a more detailed, fully sequential memory of very recent 
experience. This is especially true for skills such as turn-taking, where the recent 
history of relationships between one's own and another's actions must be attended 
and responded to. In IHA, while the experience metric space preserves the ordering of 
experiences (so that rewards over future horizons may be computed), there is not a 
mechanism to recall the most recent experience, only the most similar. Additionally, 
experiences aggregate data over a window of time, eliminating potentially useful 
fine-grained information about changes in sensor values. The short term memory 
preserves temporal information within selected sensorimotor data over a fixed timespan of the recent history. Because the short term memory captures relevant changes in sensor data in the recent history of experience and an experience accumulates sensor data, it is reasonable for the length of the short term memory to be of a longer duration than the experience length, probably several times its length. The effect of the relationship between experience length and the duration of the short term memory will be experimentally investigated in Section~\ref{sec:results}.

This additional temporal extension of data directly influencing action selection is especially important for guiding social interactions as it allows rewards to be assigned based on these histories of interaction, rather than just the instantaneous state of the interaction that the robot is currently experiencing. History-based engagement scores are included as variables in the vector that defines the experience space. Multiple engagement scores for the memory module may be defined to process specific relationships within the data. These scores are computed over the duration of a sliding window of past data values with a length defined as a parameter of the system as the experience length is. This places a performance measure of the recent history of the robot's behavior within an experience, making experiences involving similar immediate contexts that are part of either successful or unsuccessful interaction distinguishable to the system.


\subsection{Reward}

Of the possible sources of reward arising from internal motivations discussed in Section~\ref{sec_socialdev}, three were selected for the interaction experiment described in this article. 
The reward used by the system was the sum of engagement scores related to the various forms of social feedback used. One, visual attention from the interaction partner, was based on immediate perceptual information. The other two sources of motivation were based on turn-taking performance, and the short-term memory module was used to compute their engagement scores. 

 The two turn-taking behaviors assessed were peek-a-boo and drumming. Though engagement was calculated differently for each turn-taking task, the criteria for determining the quality of interaction was the same. In both cases, the criteria used to compute the engagement score was
\begin{itemize}
\item reciprocity - \emph{Are both the robot and the human playing this game?}
\item synchronization - \emph{Is each of them waiting for the other to take their turn?}
\end{itemize}
The details of how the engagement scores were computed for each source are described below.

\subsubsection{Visual Attention Score}
 
 Visual attention was detected using the gaze direction coordinates reported by the gaze tracker and the result of running the face detection algorithm on the gaze tracker's scene camera. The bounding box for the robot's face was found in the image. If the human's gaze direction (also reported in pixel coordinates of the gaze tracker's scene camera) fell within this bounding box, visual attention was set to a value of one for that timestep. Otherwise it was set to zero.
 
 \begin{figure}[h]
\centering
\includegraphics[width=\columnwidth]{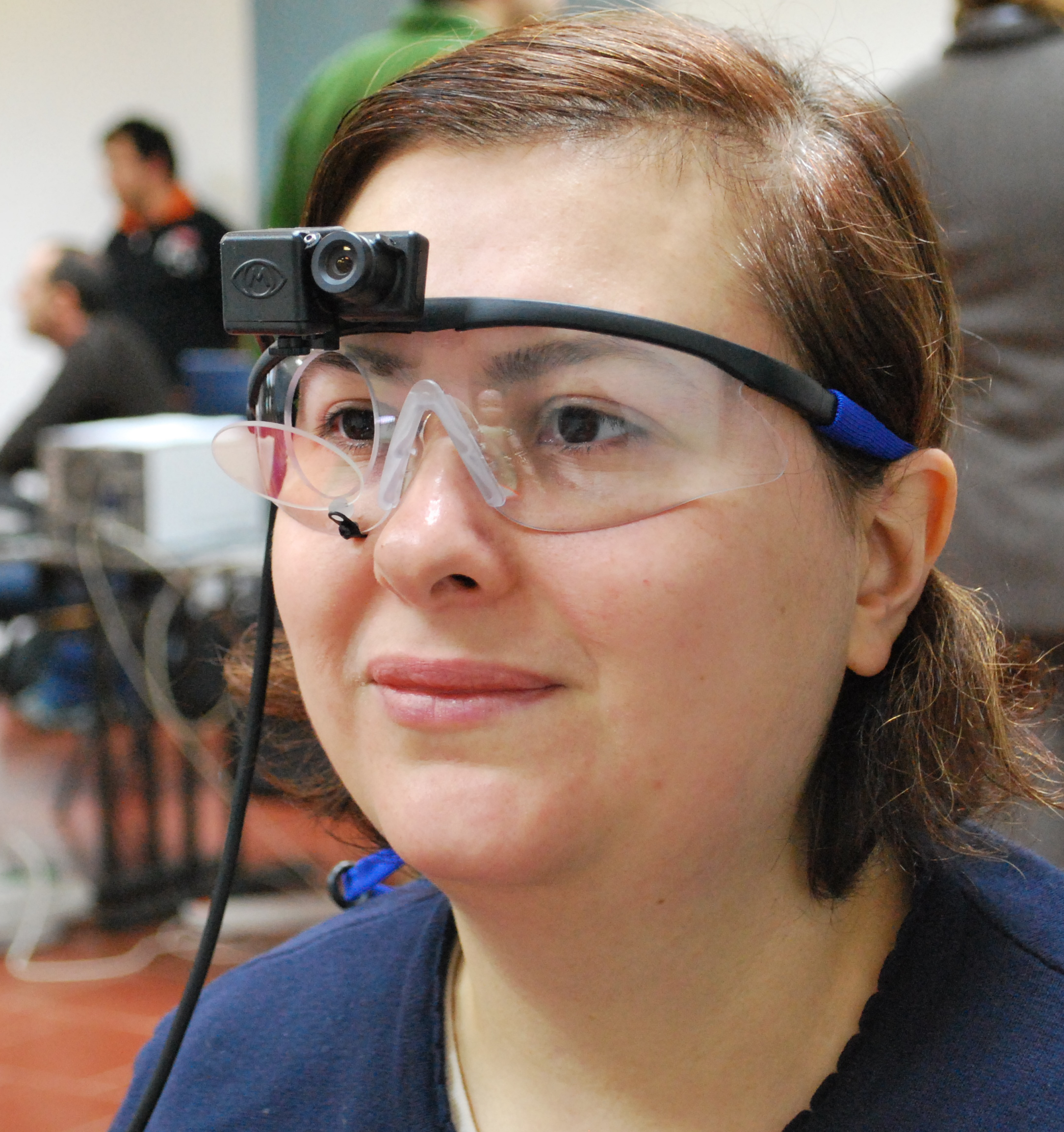}
\caption{The human interaction partner's visual attention toward the robot is measured using a gaze tracking system.}
\label{fig_gazetracker}
\end{figure}

\subsubsection{Turn-taking Scores}

Both of the turn-taking tasks used knowledge of the \emph{characteristic action} for the task to compare the behavior of the human and the robot. The characteristic action was the action in a sequence making up a behavior that was necessary in order for the behavior to be recognized as being related to a task. For peek-a-boo, the characteristic action was ``hide-face" and for drumming it was ``hit-drum". Both tasks also computed their engagement scores over the sensor history of the short term memory module, whose duration was a run-time parameter of the EIHA system. The details of the engagement score function for each task are described below. Pseudocode for the computation of the scores is shown in Figure~\ref{fig_reward}.

For peek-a-boo, at each timestep, it was determined whether the human and robot were hiding, and then whether only the human was hiding at that timestep. These variables were kept in the short term memory. Positive engagement was given when the human had been hiding for a certain length of time (between 0.5 and 2.5 seconds) while the robot is not executing the hide action. This duration of time represents the expected length of time that a person would hide their face during peek-a-boo. Shorter losses of the face are likely caused by transient failures in face detection. Longer losses are likely to be caused by a person being absent or having their face (and therefore attention) directed away from the robot. When the human's hiding action fell within this range of time, the reward was computed as the normalized sum of the durations of robot and the human's hide actions over the memory length.

 The case were the human is hiding and the robot is not was used to assign an engagement value because in the reverse scenario (robot hiding and human not hiding) the robot partially obscures its view of the interaction partner with its hands during its hide action, causing face detection failures that make it difficult to judge whether or not the human is actually hiding. 
 
For the drumming task, the variables of interest were whether the robot and human were drumming and whether they were both drumming at the same timestep. The robot was determined to be drumming only when it was currently selecting the characteristic action (so not while it was starting or ending a turn of drumming). The human was determined to be drumming when the audio analyser had detected a beat in the e-drum input within the last timestep. If the robot chose the hide action while the human was drumming, it received a fixed size negative engagement score reflecting that in this context that action appeared inattentive to the human. Also, the robot could only receive positive engagement for drumming when the human had been drumming within the history of the short term memory's length. This  constraint limits the engagement score to turn-taking interactions, rather than allowing the robot to increase its score simply by drumming at any time. The engagement score is a normalized sum of the history of the human and robot's drumming, penalized by the sum of the timesteps when both were drumming at the same time rather than synchronizing their turn-taking properly. Note that unlike in the peek-a-boo task, engagement may be negative. In this task, poor synchronization can be detected and penalized (unlike in peek-a-boo, where occlusion of the vision system during hiding may limit perception of the human's actions).

\begin{figure}[t]
\framebox[\columnwidth]{
\begin{minipage}[t]{\columnwidth}
\setlength{\parindent}{4mm}  
 \textbf{System parameters:} \\
 \indent $mem\_length$ = duration of short term memory module\\ 
 \indent sliding window (secs)\\
  \indent $min\_time$ = shortest face tracking loss assumed to be \\
  \indent human hiding face (secs)\\
    \indent $max\_time$ = longest face tracking loss assumed to be \\
    \indent human hiding face(secs)\\
    \indent $resolution$ = incoming data readings per secord\\

\textbf{Compute at each timestep:} \\ 
 \indent $human\_hide$ = no face found in robot eye camera  \\ 
 \indent $robot\_hide$ =  robot action is "hide-face" \\
 \indent $human\_only\_hide = human\_hide$  and not($robot\_hide$)  \\ 
  \indent $robot\_drum$ =  robot action is "drum-hit" \\
  \indent $human\_drum$ = beat detected from e-drum input \\
  \indent $both\_drum = human\_drum$ and $robot\_drum$  \\
  \indent Keep a running sum of the above variables over \\
 \indent their past $mem\_length$ values \\

 \textbf{Calculate peek-a-boo engagement:} \\ 
 \indent $hide\_score=0$ \\ 
 \indent if $ ((\sum{human\_only\_hide} < (resolution*max\_time))$\\ 
 \indent \indent and\\ 
\indent \indent $(\sum{human\_only\_hide}>(resolution*min\_time)))$ \\
     \indent \indent $hide\_score = \\ 
     \indent \indent \indent (\sum{human\_only\_hide}+\sum{robot\_hide})/$ \\
     \indent \indent \indent $(resolution*mem\_length) $\\ \\

 \textbf{Calculate drumming engagement:} \\ 
\indent $drum\_score = 0$ \\
 \indent if $(robot\_hide)$ \\
  \indent \indent if $(human\_drum)$ \\
\indent \indent \indent $drum\_score = -0.5$ \\
\indent else \\
\indent \indent if $(\sum{human\_drum} > 0)$ \\
\indent \indent \indent $drum\_score = (0.5*(\sum{robot\_drum}$ \\
\indent \indent \indent \indent $+ \sum{human\_drum})- \sum{both\_drum})/$ \\
\indent \indent \indent \indent $(resolution*mem\_length)$ \\
   \end{minipage}
   }
 \caption{Definitions of the memory-based engagement scores for peek-a-boo and drumming.}
\label{fig_reward}
\end{figure}

\subsection{Switching between behaviors}
    
      One important feature of this system is the ability to switch between learned behaviors.\footnote{For an example of the robot switching between learned behaviors, see the video documentation of the system at \url{http://eris.liralab.it/misc/icubvideos/ihaNew_short2_web.mov}. In this video, there are also some examples of the robot initiating attempts at switching between interaction games. In particular the robot need not necessarily always play the role of follower, but could start an interaction game of a particular kind itself. }
       Both interactions (drumming and peek-a-boo) can be learned during a single episode of extended interaction with the human teacher. Once learned, the robot can change between the two learned behaviors to maximize their reward based on social feedback from the human. For example, if the robot were playing peek-a-boo with the human, and the human quit playing and attending to the robot, the robot would eventually stop repeating the peek-a-boo behavior and start to explore new action sequences, including possibly initiating the drumming behavior. If the human responded by drumming back, the robot would continue the drumming interaction. 
      
      At first when the human quits playing, the robot's most similar experiences from the experience metric space are experiences from the history of reciprocal, high-reward turn-taking interaction with the human. As the human fails to respond, either new experiences of lower reward for the hide-face action (and thus the peek-a-boo sequence) are created or the reward associated with that action are revised downward for the recalled experience (whether experiences are added to the experience space or existing experiences are revised depends on current sensor values and on system parameters of EIHA). Due to the lower reward for the characteristic action of peek-a-boo, the probabilistic action selection rule is more likely to explore other actions. Note that any action selected (such as waving) can be reinforced using visual attention, so it is possible for the robot to learn behaviors during interaction other than drumming and peek-a-boo. When the ``start drum" action is selected, the robot will recall experiences from its recent drumming interaction with the human, making it more probable that high-reward actions associated with this history of interaction will be chosen. If the human responds to the drumming by drumming in turn, the learned drumming behavior is further reinforced.

\section{Experiment}
\label{sec:exp}

\begin{figure}[h]
\centering
\includegraphics[width=\columnwidth]{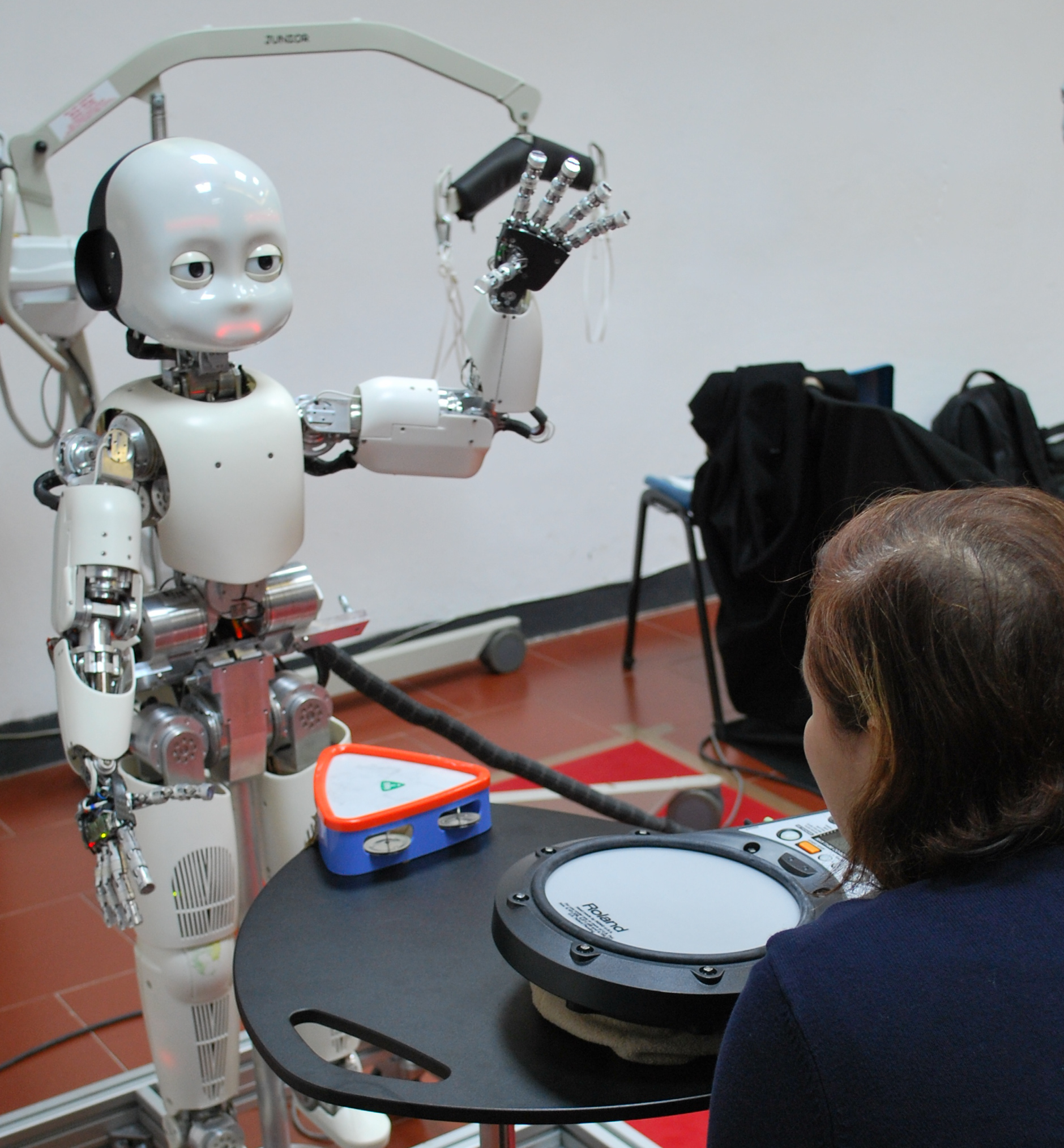}
\caption{The iCub interacts with the human interaction partner using gestures and drumming.}
\label{fig_exp}
\end{figure}

In order to evaluate the system, we need to demonstrate whether and under what conditions EIHA can support the learning of (and subsequent engagement in) turn-taking, demonstrated using the selected activities of drumming and peek-a-boo. The experimental setup was that of a face-to-face interaction with a human teacher/playmate. The human sat approximately 1 meter away from the iCub with a small table between them on which was placed the e-drum that the human used for drumming and the toy drum that the iCub used. An example of the setup is shown in Figure~\ref{fig_exp}. During the experiment, the human wore a wearable gaze tracker like the one shown in Figure~\ref{fig_gazetracker}. 

The impact of the short term memory module's addition in EIHA was experimentally investigated by allowing the robot to play the game with a human for multiple trials and examining the outcome in terms of which behaviors were successfully learned. Three conditions were compared: short term memory (STM), truncated STM, and no STM.  In the no STM condition, the short term memory module is not used by the system. The only source of reinforcement in this condition was visual attention through gaze because the other sources of reward rely on engagement scores calculated using information from the short term memory module. Note that in this condition EIHA operates in the same manner as IHA, with the only difference between the systems being the extended sensor input in this implementation. In the STM condition, the robot received reinforcement from both visual attention and the turn-taking engagement scores from the short term memory module. In this condition, the duration of the interaction history used by the short term memory module was 4 seconds, a length of time chosen empirically based on the duration of the relevant turn-taking behaviors. This length of time should typically capture at least one transition between the robot's and human's turn during well-coordinated turn-taking for both of the behaviors  to be learned. In the truncated STM condition, the short term memory module computed rewards based only on a 1 second history of interaction data, a length of time expected not to be long enough to be able to capture the turn-taking dynamics of both behaviors.  The experience length used by this system was 2 seconds in duration at a 10 Hz sensor rate. The parameter settings for EIHA were tuned in the earlier implementation of peek-a-boo learning on the iCub with the original version of IHA.

The robot performed 5 game trials per condition. Each trial lasted until either both turn-taking behaviors had been learned successfully (the criteria by which this was determined is described in the next section) or 10 minutes had elapsed.  All of the trials were performed with a single trained interaction partner. The experiment was conducted in this manner (rather than having the robot interact with several naive users) in order to focus on differences due to task complexity rather than individual interaction styles. Also, the use of an experienced interaction partner eliminated the possibility of training effects over the course of the trials. 

\section{Results}
\label{sec:results}

One way of measuring the capabilities of the system under the different STM models is to look at whether turn-taking behaviors were learned during the trials. The criteria used to determine whether a turn-taking  behavior is learned is the occurrence of three consecutive robot-human turns of that behavior. The ``consecutive" requirement means that if the robot introduces additional actions that are not part of the desired behavior sequence into a turn-taking interaction, the turn-taking is considered to not have been learned successfully and previous turns are not counted. This number was chosen experimentally by observing that once the robot has had a behavior sequence reinforced to the point where it can engage in three consecutive turns, it will typically continue turn-taking with the human as long as it continues to receive positive reinforcement. It is also able to switch back to the learned behavior later on in an interaction once this criteria has been met. The turn-taking was evaluated by manually coding video recorded of the experimental interactions with the robot. This method was used rather than relying on data logged from the robot in order to ensure that the human's turn-taking actions were correctly identified.

There were two criteria used to evaluate the learning performance of the different models. First, whether a model could support the learning of a behavior within the 10 minute trial length was considered. In cases where different models were capable of learning a behavior, differences in performance were evaluated in terms of the amount of time it took for the system to learn the behavior after the first time the robot was observed to perform the \emph{characteristic action} for a behavior.

\subsection{Learning success}

\begin{figure}[h]
\centering
\includegraphics[width=\columnwidth]{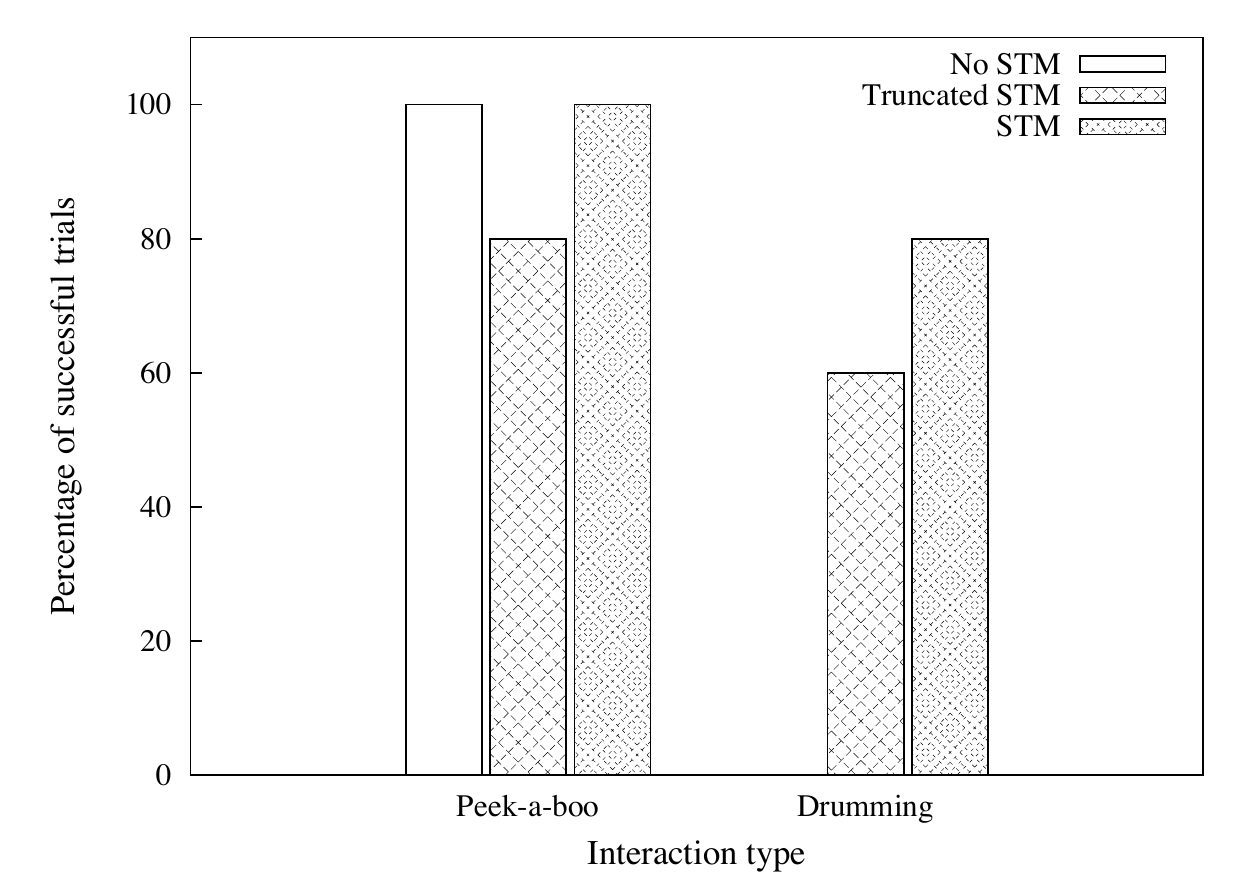}
\caption{The percentage of trials which resulted in at least 3 successful rounds of turn-taking of either desired turn-taking behavior.}
\label{fig_success}
\end{figure}

The difference in the capability of the models to learn the behaviors within the given trial time was compared. The percentages of the trials that resulted in the successful learning of each behavior are shown in Figure~\ref{fig_success}. It can be seen that all of the models successfully learned the peek-a-boo behavior during almost all of the trials. The differences in learning success for the drumming behavior were greater. None of the models successfully learned drumming in every trial, and the no STM model did not learn it during any trial. Fischer's exact test was used to evaluate the significance of the differences between the proportions of successes and failures of the STM model and the other models (see Table~\ref{table_fischer}). Based on the number of successes versus failures alone, the no STM model's results showed a statistically significant difference for the drumming interaction (p-value = 0.02). This is a strong indication that drumming was the more difficult to learn of the two forms of turn-taking, and that the short term memory module played a role in the system's ability to learn it.

\begin{table}[h]
\caption{Results of Fischer's exact test comparing the proportion of successes and failures for the memory model versus the other models. }
\label{table_fischer}
\centering
\begin{tabular}{|l||c|}
\hline
\multicolumn{2}{|c|}{Peek-a-boo} \\ \hline
Model & p-value \\ \hline
Truncated STM & 0.5\\ \hline
No STM & 1.0\\ \hline
\multicolumn{2}{|c|}{Drumming} \\ \hline
Model & p-value \\ \hline
Truncated STM & 0.42\\ \hline
No STM & 0.02*\\ \hline
\end{tabular}
\end{table}

\subsection{Learning time}

\begin{figure}[h]
\centering
\includegraphics[width=\columnwidth]{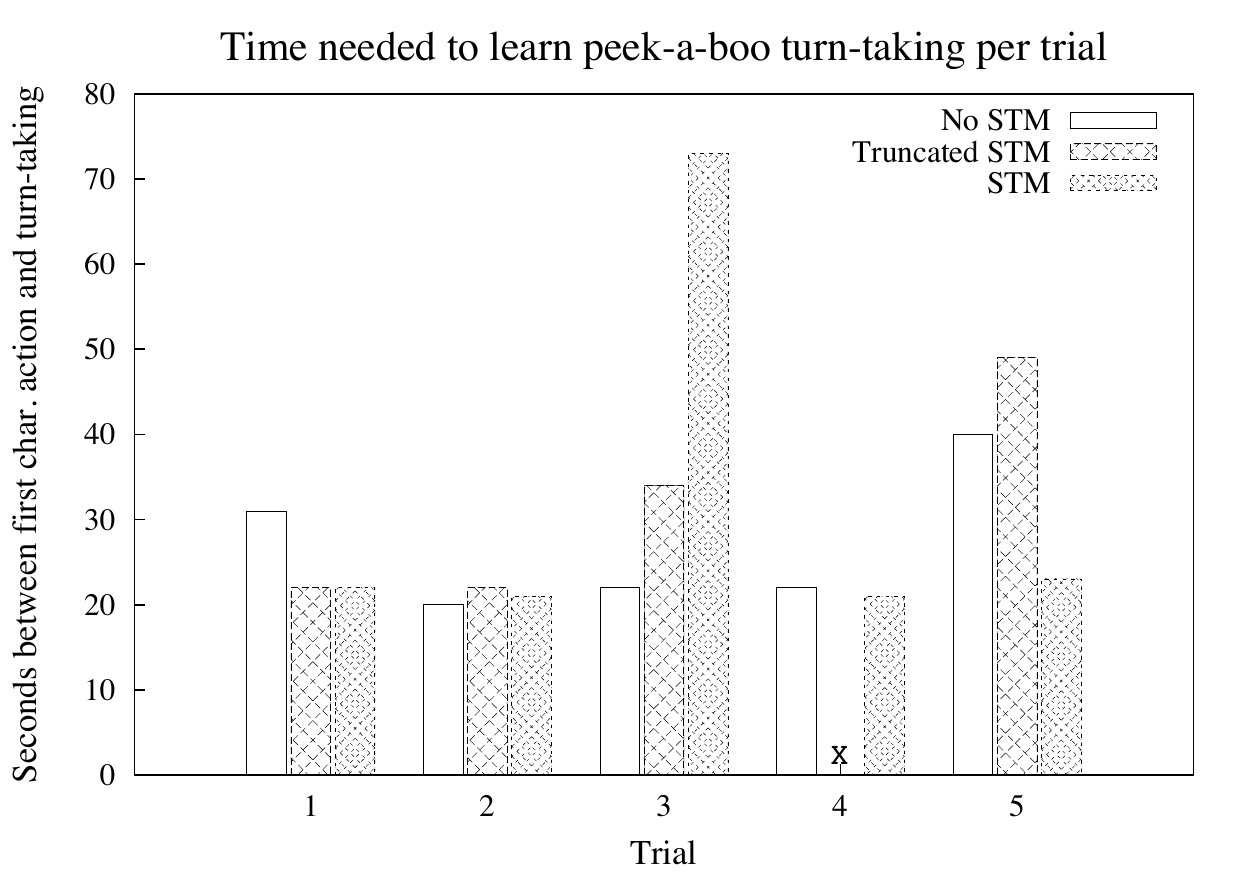}
\caption{The time in seconds between the first instance of the peek-a-boo behavior and the completion of 3 rounds of successful peek-a-boo turn-taking for each trial. X's indicate that the behavior was not learned before the time limit for that trial.}
\label{fig_peek-a-boo}
\end{figure}

\begin{figure}[h]
\centering
\includegraphics[width=\columnwidth]{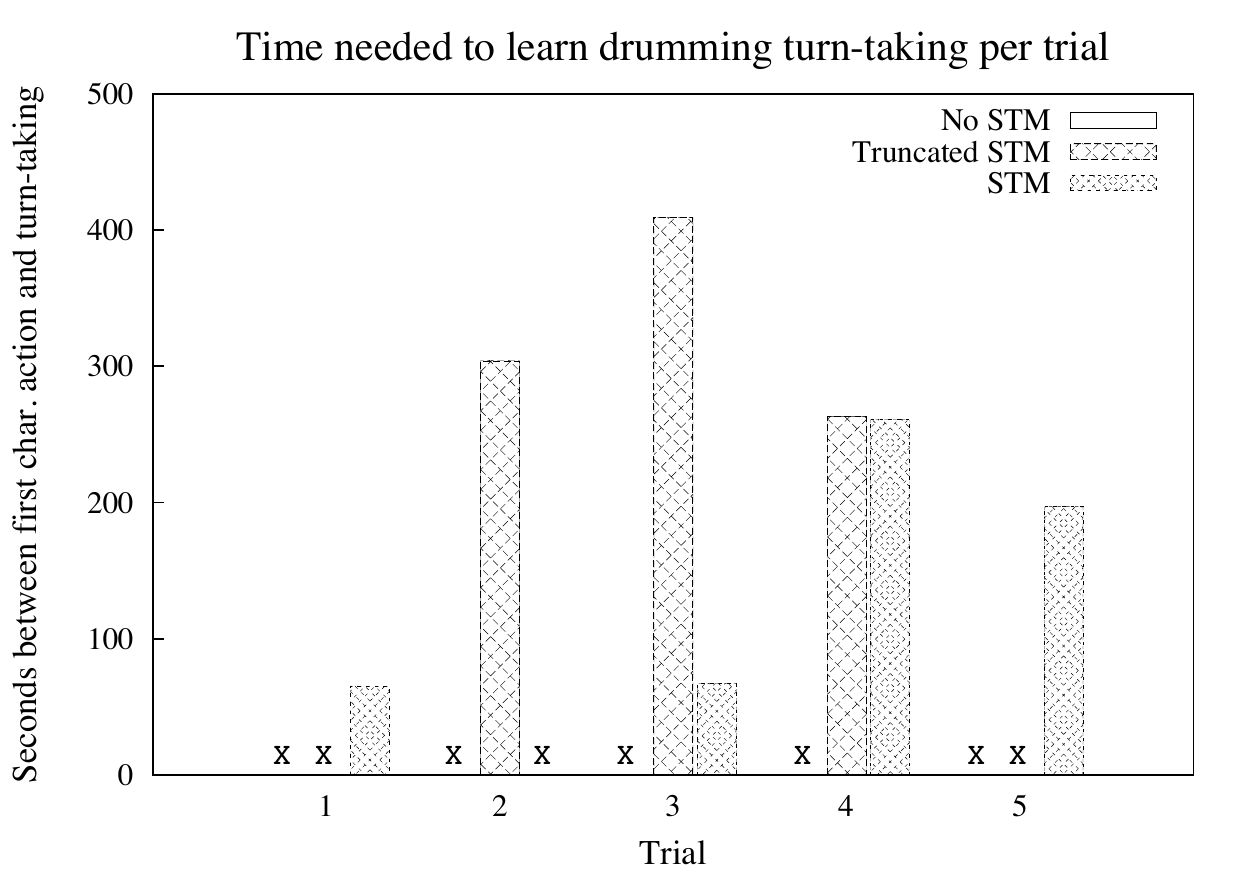}
\caption{The time in seconds between the first instance of the drumming behavior and the completion of 3 rounds of successful drumming turn-taking for each trial. X's indicate that the behavior was not learned before the time limit for that trial.}
\label{fig_drum}
\end{figure}

In addition to differences in the number of successful trials, it was also expected that there might be differences in the amount of time it took the system to have the desired behaviors sufficiently reinforced to learn turn-taking. The metric used to measure the time it took to acquire a turn-taking behavior was as follows: the start time was determined to be the first time the robot performed the \emph{characteristic action} for a mode of turn-taking interaction and the end time was determined to be the completion of the third of three consecutive turns of that interaction by the robot. As previously described, for the peek-a-boo interaction, the characteristic action was ``hide-face". For the drumming interaction, the characteristic action was ``drum-hit". The times for the trials are shown in Figures~\ref{fig_peek-a-boo}\&\ref{fig_drum}. Times for unsuccessful trials are shown as missing bars (marked with an ``X") in the graphs. Note that of the 5 trials, the robot failed to learn drumming turn-taking once under the STM model condition. The robot failed to learn peek-a-boo once and drumming twice under the truncated STM condition. Under the no STM condition, the robot did not learn drumming during any of the  trials.

In order to determine whether the difference between the average times for the models to learn a turn-taking behavior were statistically significant, randomized permutation tests using 1000 samples were performed. This test was chosen because of its ability to handle the small sample sizes and uneven number of samples (due to the unsuccessful trials) present in the data. The results are presented in Table~\ref{table_differences}. No significant differences were found between the average learning times for the successful peek-a-boo trials. The difference between the the average times to learn drumming for the successful trials of the STM and truncated STM models is moderately significant.\footnote{The randomized permutation test is not an exact test, so the 0.05 significance level is approximate.} This statistical support, combined with the difference in the number of successful trials (4 for the STM model and 3 for the truncated STM model), suggests that there is a real difference in the system's ability to learn the drumming interaction depending on the length of the short term memory used. The learning time for the no STM model couldn't be compared for drumming because there were no successful trials, demonstrating that the STM module was a necessary component of EIHA in order to learn this interaction.

\begin{table}[h]
\caption{Difference between average times for memory model and other models for behavior acquisition with significance results from randomized permutation tests. }
\label{table_differences}
\centering
\begin{tabular}{|l||c|c|}
\hline
\multicolumn{3}{|c|}{Peek-a-boo} \\ \hline
Model & Avg diff (s) & p-value \\ \hline
Truncated STM & -0.25 & 0.97\\ \hline
No STM &-5.0 & 0.85\\ \hline
\multicolumn{3}{|c|}{Drumming} \\ \hline
Model & Avg diff (s) & p-value \\ \hline
Truncated STM & 177.8 & 0.05* \\ \hline
No STM & -- & --\\ \hline
\end{tabular}
\end{table}


\subsection{Discussion}

These results suggest that a short term memory of interaction is beneficial for the learning of some types of turn-taking in EIHA, but may not have an impact on others. In order to understand why peek-a-boo could be learned quickly by all of the models while drumming could not, it is useful to look at the characteristics of the two behaviors. Peek-a-boo was a simpler interaction overall, requiring the learning of shorter action sequences with less variability in the durations of turns. 

The no memory model could not learn the drumming interaction at all, showing that visual attention feedback alone was not sufficient for the human teacher to shape this behavior. The results also demonstrate that the length of the short term memory had an impact on the amount of time it took to learn the drumming turn-taking, with the truncated STM model requiring a longer time to learn the behavior than the memory model. This is likely because the shorter duration of the truncated memory contained less information about the transitions in turn-taking between robot and human (and vice-versa), making it harder to distinguish examples of successful turn-taking by their associated reward, which is partially determined by engagement scores based on information in the short term memory.

In Mirza's work on peek-a-boo learning with the original version of IHA, the learning success was found to be dependent upon the experience length~\cite{mirza07grounded}. It was hypothesized that the duration of an experience should be related to the duration of an action performed by the robot in order to support effective learning. The short term memory module provides a way of relating the recent history of the robot's actions to those of its interaction partner. It allows reward values to be based on changes and relationships in the sensor data over time in a fine-grained way. We see in this experiment that the duration of the short term memory also has an impact on learning success. The short term memory should be long enough to capture the dynamics of the actions of both participants in an interaction, and therefore is most effective when it is longer than the experience length and also based on the duration of the human partner's actions.

\section{Enactive Development: Ontogeny without Representation}
\label{sec:enactive}

This work is can be viewed as compatible with an enactive approach to cognitive systems 
that  has been suggested and described by philosophers such as Varela, Hutto and No\"{e} and used as a guiding principle for the implementation of working systems by artificial intelligence researchers such as Vernon,  Dautenhahn and Nehaniv
~\cite{varela91,dautenhahn96remembering,Quick99,nehaniv2000,mirza07grounded,mirza08developing,vernon07,vernon10,hutto12,noe}. The enactive approach focuses on behavior and interaction and has been identified as a promising framework for the development of cognitive systems.  For the system to acquire and select behaviours in interaction, there is no need to assume the existence of static entities with properties characterizable by truth-functional propositional logic and semantics. Although robot, environment and interaction partner clearly mutually influence and shape each others' behaviors, we follow Brooks 
 in that there is no representation or model of the world, nor of the self, nor entities in it (``The world is its own best model'') ~\cite{brooks91intelligence}, however, unlike classical behavior-based systems, our robot learns new behaviors to select from. 
 
 In an enactive cognitive system the ontogeny of cognitive capacities depends on the processes of sensorimotor and social interaction and the capacity of the cognitive architecture to allow behavior to be shaped by experience (operationally defined as the temporally extended flow of values over sensorimotor  and internal variables) and feedback in terms of degree of social engagment,  rather than on modeling, symbols or logics.

 Different sequences of actions are acquired  as regular behaviors during interaction.  The meaningfulness of the interaction for the human when the robot engages with the human in more or less complex turn-taking behaviors and the switching between them are habits of activity dynamically acquired.  These behavioral trajectories can be viewed as  ``paths laid down in walking''~\cite{varela91} in a space of possibilities for social interaction that shape habits and scaffold future development.  This type of  interactive history architecture exhibits the capacity for anticipatory prospection \cite{mirzaALIFE} and constructing its own behaviors in interactive development with its environment \cite{mirza08developing}.  

Switching between acquired behaviors has been demonstrated here, but scaffolding of more complex, new behaviors that build on simpler acquired ones remains a near-term goal for this type of approach which the architecture should with little or no extension be able to support. By following exploration in a zone of proximal development (ZPD) (Vygostsky \cite{Vygotsky78}), variants and new actions could be selected in the context of interactions based on already acquired behaviors. Arguably, switching  behaviors appropriately  in response to social engagement cues is already a primitive case of this.  

The crucial role of timing and relating different scales of remembering at different scales of temporally extended experience and behavior  as hinted here and in previous work by Mirza~\cite{mirza07grounded}. Short term memory (as studied here) would seem to have an impact on developmental capacity. Further study of its role in development as well as of types of longer term narrative memory would be fruitful directions for future research.   
The interactions are ``meaningful" in the sense of Wittgenstein \cite{WittgensteinPI} in that behaviors are used in structuring interaction games via enaction of the acquired behavioral sequences appropriately in context leading to rewarding social engagement  (related to Peircean semiosis \cite{nehaniv2000}). Imitation and timing play important communicative roles in structuring the contingent interactions achieved here, as they do in the case of human cognitive development \cite{Trevarthen79,Nadel99}.

\section{Future Work}

The role of learning in EIHA is currently restricted to the learning of the 
experience space and the associations between experiences and rewards in order to 
find effective behavior sequences. But there are opportunities to extend the learning capabilities of this system in order to expand the number and types of turn-taking interactions that the system could learn by making the discovery of turn-taking relationships in the datastream automatic. While the relationships between sensors monitored for feedback about  turn-taking were predefined in this case, one could instead use statistical methods to discover  which sensor channels are associated and predictive of one another. One possibility would be to explore the use of the cross-modal experience metric proposed by Nehaniv, Mirza and Olsson~\cite{nehaniv07development}. Interpersonal maps, as described by Hafner and Kaplan, are another way of identifying these relationships using information metrics~\cite{hafner06interpersonal}. This would allow for task-specific turn-taking cues to be discovered, as well as general task-invariant cues. 

 Currently, EIHA finds the distance between individual experiences. But it may be useful to base recall on a longer temporal horizon of traces of past experiences. The topology of the experience metric space could be used to make generalizations about experiences over time, using the clustering of experiences to identify behaviors or interactions. The ability to make aggregate representations of these clusters that capture their fundamental properties could possibly allow for more powerful predictions based on current experience, opening up the potential of anticipating the actions of others as well as responding to them.

While gaze is used in this system as a form of social feedback, the robot has no active gaze behavior. Gaze plays an important part in regulating face-to-face turn-taking. It would be useful to support turn taking interactions with appropriate gaze behavior on the robot's part, either by learning gaze behaviors or by making use of a separate gaze controller that is capable of producing socially appropriate gaze in response to a human partner.

\section{Conclusion}

An enactive cognitive architecture based on experiential histories of interaction for social behavior acquistion that uses reward based on a short term memory of interaction was presented. An implementation that allowed the acquisition of and selection between behaviours for interaction games using reinforcement from social engagement cures (visual attention and behavior-specific measures of turn-taking success) was described. The architecture provides the developing agent with prospection and behaviour ontogeny based on an operationalization of the temporal extended experience, and compared current experience to prior experiences to guide behaviour and action selection. Moreover, since it chooses actions based on prior experience or at times randomly, the robot may play the role of initiator of the particular type of turn-taking game in interaction with an human partner. The significance of the short term memory and the importance of using a short term memory length that is capable of capturing characteristics of the desired interaction behaviors in such an architecutre was demonstrated experimentally, showing that the short term memory had a beneficial effect both on the system's ability to acquire some behaviours and the amount of time required to acquire turn-taking behaviors. The capacity not only to learn new behaviours, but to actively switch between them depending on the social cues in the context of interaction with a human have also been demonstrated.


%



\section*{Acknowledgment}

This research was mainly conducted within the EU Integrated Project RobotCub (Robotic Open-architecture Technology for Cognition, Understanding, and Behaviours) and was funded by the European Commission through the E5 Unit (Cognition) of FP6-IST under Contract FP6-004370. The work was also partially funded also from ITALK:  Integration and Transfer of Action and Language Knowledge in Robots under Contract FP7-214668. These sources of support are gratefully acknowledged.


\ifCLASSOPTIONcaptionsoff
  \newpage
\fi



\bibliographystyle{IEEEtran}
\bibliography{eiha}
\end{document}